\documentclass[lettersize,journal]{IEEEtran}
\pdfoutput=1
\usepackage{amsmath, amsfonts, amssymb, amscd, mathtools, physics}
\usepackage{algorithmic}
\usepackage{array}
\usepackage[caption=false,font=normalsize,labelfont=sf,textfont=sf]{subfig}
\usepackage{textcomp}
\usepackage{stfloats}
\usepackage{url}
\usepackage{verbatim}
\usepackage{graphicx}
\def\BibTeX{{\rm B\kern-.05em{\sc i\kern-.025em b}\kern-.08em
    T\kern-.1667em\lower.7ex\hbox{E}\kern-.125emX}}
\usepackage{balance}

\usepackage[utf8]{inputenc} % allow utf-8 input
\usepackage[T1]{fontenc}    % use 8-bit T1 fonts
\usepackage{caption} 
\usepackage[hidelinks]{hyperref}       % hyperlinks
\usepackage{url}            % simple URL typesetting
\usepackage{booktabs}       % professional-quality tables
\usepackage{nicefrac}       % compact symbols for 1/2, etc.
\usepackage{microtype}      % microtypography
\usepackage{xcolor}         % colors

\usepackage{dsfont}
\usepackage{orcidlink}
\usepackage{color}
\usepackage{cleveref}

\begin{document}

\title{Beyond Size and Class Balance: Alpha as a\\New Dataset Quality Metric for Deep Learning}
\date{\today}

\author{
    Josiah~Couch\orcidlink{0000-0002-7416-5858}, %
    Rima~Arnaout\orcidlink{0000-0002-7134-0040}, %
    and~Ramy~Arnaout\orcidlink{0000-0001-6955-9310}%
        % <-this % stops a space
\thanks{%Manuscript received ... 
This work was supported by the Gordon and Betty Moore Foundation and by the NIH under grants R01HL150394, R01HL150394-SI, R01AI148747, and R01AI148747-SI.}% <-this % stops a space
\thanks{Josiah Couch is with the Department of Pathology at Beth Israel Deaconess Medical Center (BIDMC), Boston, MA 02215. Rima Arnaout (rima.arnaout@ucsf.edu) is with the Department of Medicine, the Bakar Institute for Computational Health Sciences, and the Center for Intelligent Imaging at the University of California San Francisco, San Francisco, CA 94143. Ramy Arnaout (email: rarnaout@bidmc.harvard.edu) is with the Department of Pathology and the Division of Clinical Informatics, Department of Medicine, BIDMC and with Harvard Medical School, Boston, MA 02215.}% <-this % stops a space
%\thanks{Digital Object Identifier ...}
}

\maketitle
\begin{abstract}

In deep learning, achieving high performance on image classification tasks requires diverse training sets. However, the current best practice---maximizing dataset size and class balance—does not guarantee dataset diversity. We hypothesized that, for a given model architecture, model performance can be improved by maximizing diversity more directly. To test this hypothesis, we introduce a comprehensive framework of diversity measures from ecology that generalizes familiar quantities like Shannon entropy by accounting for similarities among images. (Size and class balance emerge as special cases.) Analyzing thousands of subsets from seven medical datasets showed that the best correlates of performance were not size or class balance but $A$---``big alpha''---a set of generalized entropy measures interpreted as the effective number of image-class pairs in the dataset, after accounting for image similarities. One of these, $A_0$, explained 67\% of the variance in balanced accuracy, vs. 54\% for class balance and just 39\% for size. The best pair of measures was size-plus-$A_1$ (79\%), which outperformed size-plus-class-balance (74\%). Subsets with the largest $A_0$ performed up to 16\% better than those with the largest size (median improvement, 8\%). We propose maximizing $A$ as a way to improve deep learning performance in medical imaging.

\end{abstract}


\tableofcontents

\section{Introduction}\label{sec:intro}

For deep learning to reach its potential in medicine and other fields, models must be data-efficient, highly performant, and generalizable \cite{spector2023, sachdeva2024, dey2023b}. Investigators have long appreciated that models generally perform better on datasets that are more diverse. As a working definition, a diverse dataset is one whose elements are sufficiently representative of the intended real-world application for a model to be able to learn class invariants that generalize beyond the dataset. Perhaps the most commonly used strategy to promote diversity is to maximize dataset size and class balance: the larger the dataset, the more likely its elements will fill out the space of real-world possibilities; the more class-balanced, the more elements per class.

Indeed, size and class balance have long been the two primary heuristics of dataset quality, which investigators have sought to maximize for applications including (but not limited to) image classification, the focus of the present work, since well before the explosion of deep learning over the past 20 years \cite{simard1991, yaeger1996, shorten2019, lecun1995, buda2018, athalye2023a}.

Yet from this working definition, it should be clear that size and class balance alone will not generally provide a complete description of dataset diversity, since they only indicate how many images there are (overall and in each class); they do not indicate how well these elements fill out the space of possibilities, which is related to how similar or different the images in the dataset or class are to each other. Given that constructing large, class-balanced datasets is often challenging---especially in medicine, where acquiring and annotating high-quality images can be expensive and/or labor-intensive---the conventional approach of seeking to maximize size and class balance might not always be the most efficient strategy for achieving high-quality datasets. To improve on current practice, one approach is to ask whether there exist measures of dataset diversity that account for similarities and differences among the elements in the dataset, and if so, whether these measures might outperform size and class balance as correlates of model performance (for a given model architecture).

The present work demonstrates that the answer to both questions is “yes,” suggesting a new potential avenue for improving machine learning performance by optimizing measures of dataset quality that capture similarities and differences among dataset elements. Specifically, we identify $A$, ``big alpha,’’ and more specifically measures known as $A_{q=0}$ (``big alpha at $q$ equals 0;’’ henceforth $A_0$) and $A_1$, as the best correlates of AUC, F1 score, accuracy, and balanced accuracy, using ResNet-18 and a variety of medical-imaging datasets across a range of imaging modalities as proof of principle. $A$ is part of a comprehensive mathematical framework developed in ecology over the past decade that generalizes the R\'enyi entropies \cite{renyi1961} (of which Shannon entropy \cite{Shannon:1948zz} is a special case) to create a family of similarity-sensitive diversity measures. We refer to this framework as ``LCR,’’ after its authors' initials (Section 2) \cite{leinster2012, reeve2016, leinster2021}. Using ultrasound, X-ray, CT, histopathology, and hematology image datasets (Section 3), here we present an approach for assessing how well each diversity measure correlates with performance (Section 4), and describe the results of this approach (Section 5). We conclude with a discussion of limitations and future directions (Section 6).

\subsection{Relationship to prior work}\label{subsec:related}

There is a rich literature describing issues related to dataset composition in the context of machine learning, and an extensive literature on potential solutions. Data augmentation is a standard solution in imaging (and other) applications that was originally introduced as a way to increase dataset size, although even early works make it clear that the implicit goal was to increase diversity (an early example of size as a proxy for diversity) \cite{simard1991, yaeger1996, shorten2019, lecun1995, athalye2023a}. The value of class balance as a contributor to model performance has long been recognized and has been extensively described. Regarding evaluation of model performance, prior work has also described the danger of using ``vanilla’’ accuracy for performance evaluation when test sets are imbalanced \cite{megahed2021}, since in cases of severe imbalance a model may be able to achieve high accuracy simply by ignoring the smallest class(es); the general solution has been to evaluate performance using balanced accuracy (BACC) instead (related to the F1 score), as we do in the present work.

Investigations of dataset ``complexity,’’ by various definitions, as a determinant of model performance independent of size and class balance predate discovery of LCR's systematic diversity framework by over a decade \cite{japkowicz2000}. This includes appreciation of within-class diversity, discussed in the context of within-class imbalance and the problem of small disjuncts \cite{holte1989, japkowicz2001, jo2004}. Such work demonstrates long-standing appreciation of a world beyond size and class balance, although of course not investigated systematically according to LCR or to our knowledge in the context of modern deep learning or medical imaging. More generally, many methods for measuring dataset complexity have been proposed, some defining complexity specifically in the context of model performance (e.g. ``hardness’’) and others defining it based on dataset and/or image composition, but none using LCR \cite{ojima2021, cho2021, chen2022, paiva2022}. Of note, some of the latter methods employ entropy within images (e.g. color entropy \cite{ivanovici2020}) as opposed to across images within a dataset, as we do here; the focus of this prior work was investigating how best to estimate similarities between images, an important issue that is largely beyond the scope of the present work, as opposed to measuring the diversity of the dataset as a whole, which is our focus in the present work.

Finally, data optimization methods \cite{wan2022} such as dataset pruning,  distillation, and boosting \cite{he2023, chen2023, sundar2023, guo2004}; instance selection \cite{chinn2023a}; and to some extent active learning \cite{aggarwal2020, zhan2022} are all concerned with creating smaller datasets that achieve the same model performance as larger ones. In this sense, all are motivated by the goal of preserving dataset diversity (indirectly and/or by other names) irrespective of size and in some cases of class balance, by maximizing bespoke quantities whose potential relationships to the first-principles entropic formulation of LCR as a unifying framework are as yet uninvestigated. However, the emphasis in these prior works was how to optimize these quantities (which we speculate might prove to be correlates of LCR diversity in some fashion), whereas the goal of the present work is to determine what specific LCR diversity measure(s) best predict model performance (with the question of how best to optimize them, algorithmically or mechanistically, left for future investigation).

\section{Background}\label{sec: background}

\begin{figure}[t]
    \begin{center}
    \includegraphics[width = 0.45\textwidth]{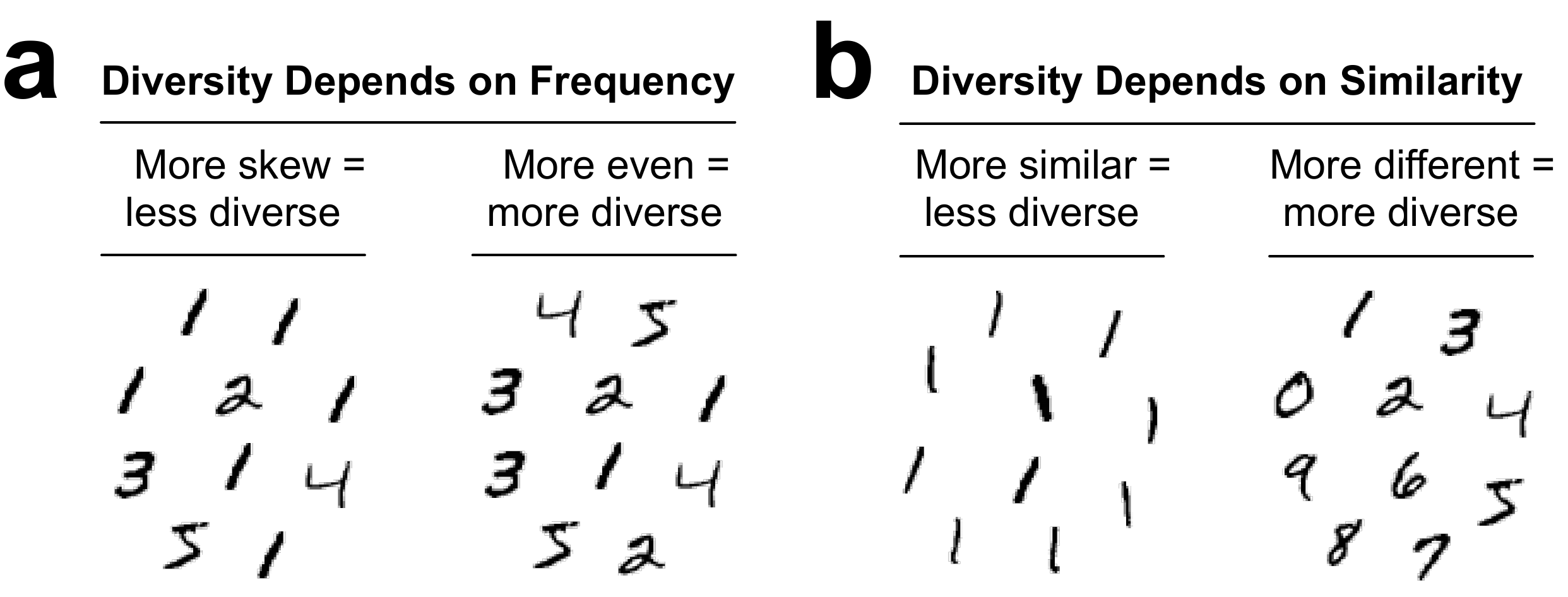}
    \caption{Sensible definitions of diversity must be sensitive to both frequency and similarity. Four same-sized datasets of 10 images each from the MNIST handwritten digits dataset \cite{mnist} are shown. The two datasets in (a) contain the same five unique images, differing only in their relative frequencies; the more balanced dataset is intuitively more diverse. The two datasets in (b) each contain 10 unique images but differ in how similar the images are to each other; the dataset with the more-different images is intuitively more diverse. See also e.g. \cite{jost2007, greylock}.}
    \label{fig: fig_diversity}
    \end{center}
\end{figure}

\subsection{Dataset diversity as a function of element frequencies and similarities} 

Motivating our approach is the observation that measures of dataset diversity make intuitive sense only if they account for both the relative frequencies of the elements in a dataset and the similarities and differences among these elements (Fig. \ref{fig: fig_diversity}) \cite{greylock}.

The effect of relative frequencies is less important in the context of machine learning datasets because elements each generally appear only once (although consecutive video frames can be quite similar \cite{arnaoutEnsemble2021}), but is worth illustrating because it factors into LCR (as the relative sizes of overlapping similarity clusters; below). Consider two datasets that each contain five unique images (Fig. \ref{fig: fig_diversity}a). In the first dataset, one image appears six times and the other four each appear only once, while in the second dataset, each image appears twice. Both datasets consist of 10 images, but the second dataset is intuitively more diverse, since the first is dominated by, and therefore to a first approximation consists of, just a single image.

To illustrate the effect of similarities and differences, consider another two datasets of 10 images, all unique, where the images in the first dataset are very similar to each other and those in the second dataset are very different from each other (Fig. \ref{fig: fig_diversity}b). Even though the datasets are identical by every purely frequency-based measure---including size, Shannon entropy, and every R\'enyi entropy---the second dataset is intuitively more diverse. Thus, measures of dataset diversity must also incorporate similarity, in addition to frequency, to align with the intuitive sense of what ``diversity’’ should mean.

\subsection{Hill's D-number framework and element frequencies}

LCR is based on Hill's diversity indices \cite{hill1973}, also known as the Hill numbers or D numbers:

\begin{eqnarray}\label{eq: Hill}
    D^q(p) &=& \left[ \sum_i p_i^q \right]^{\frac{1}{1-q}} \\
    &=& \exp\left[ H^\alpha(p) \right]
\end{eqnarray}

$D^q(p)$ are the ``(similarity-insensitive) diversities of order $q$.'' They are the exponentials of the R\'enyi entropies $H^\alpha(p)$ \cite{renyi1961} ($\alpha = q$) and the reciprocals of the generalized (power) means of order $q-1$. $p$ is a vector that represents the relative frequencies of unique elements such that $p_i$ represents the relative frequency of the $i\textsuperscript{th}$ element and $\sum_i p_i = 1$. 

The parameter $q$ down-weights the contribution of rarer elements to the total. Notably, several common statistics are simple transformations of the Hill numbers at different $q$, with size, Shannon entropy, Simpson's index, and the Berger-Parker index corresponding to $q=0$, $1$, $2$, and $\infty$, respectively \cite{jost2007}. It is useful to think of these measures as an ordered set that differ purely by $q$, i.e. as counts that increasingly ignore rarer elements. D-number forms have the advantages of self-consistent partitioning, of using the same units regardless of $q$, and of having a common interpretation as giving the ``effective number'' of elements in the dataset, discounting rarer elements proportional to $q$ \cite{jost2007}. Note that while Hill numbers account for the relative frequencies of elements in a dataset, they do not account for elements' similarities.

\subsection{The LCR framework and element similarities}

LCR \cite{leinster2012, reeve2016, leinster2021} generalizes Hill's framework by accounting for similarity: 

\begin{eqnarray}\label{eq: LCR}
    D_Z^q(p) &=& \left[ \sum_i p_i (\mathbf{Zp}_i)^{q-1} \right]^{\frac{1}{1-q}} \\
    &=& \exp\left[ H_Z^\alpha(p) \right]
\end{eqnarray}

$D_Z^q(p)$ is the similarity-sensitive diversity of order $q$. $H_Z^\alpha(p)$ is the corresponding similarity-sensitive R\'enyi entropy. The new piece relative to the Hill numbers is $Z$, the \textit{similarity matrix}, whose entries $Z_{ij}$ represent the pairwise similarity between species $i$ and species $j$. $\mathbf{Zp}_i=\sum_j Z_{ij}p_{j}$ is interpreted as how ``ordinary'' the $i\textsuperscript{th}$ unique element is, given all the other elements in the dataset; the more ``ordinariness'' in the dataset, the less diverse it is \cite{leinster2012}. Similarities $Z_{ij}$ are generally assumed to lie in the interval $[0, 1]$, with every species having a similarity of $1$ with itself. It is often also sensible to assume that $Z$ is symmetric ($Z_{ij}=Z_{ji}$). We follow these two assumptions in the present work.
 
\subsection{Dataset size and class balance as special cases of LCR}

Several quantities of interest emerge as special cases of the LCR framework. Dataset size is the special case of LCR with $Z=I$ (the identity matrix) and $q=0$. More generally, the Hill framework is the special case of LCR with $Z=I$ for any $q$, making the traditional/similarity-insensitive versions of Shannon entropy, Simpson's index, and the Berger-Parker index also special cases of LCR. By extension, $D^q_Z$ for $Z\neq I$ can be interpreted as the extended/generalized/similarity-sensitive versions of (the effective-number forms of) Shannon entropy, Simpson's index, and so on, with the details of the similarity sensitivity depending on the choice of $Z$. In particular, when $Z$ has a block structure in which $Z_{ij}=1$ when elements $i$ and $j$ are in the same labeled class and $Z_{ij}=0$ otherwise, $D^q_Z$ gives the class balance. For example, the Shannon entropy of the relative class sizes, in effective-number form, is $D^1_Z$: the effective number of classes in the dataset, accounting for their relative sizes. Thus, LCR provides a logical organizing principle for a very wide range of measures of dataset diversity, including the ones used most often in machine learning.

\subsection{Class-level and dataset-level diversities}

In LCR, each labeled class of a dataset can be assigned an $\alpha$, $\beta$, and $\gamma$ diversity. These are each defined at a given $q$; for readability we include the subscript only when necessary. $\alpha$, $\beta$, and $\gamma$ correspond roughly to the diversity within the class, the diversity of the elements in the class relative to the overall dataset, and the diversity of the class as a whole relative to the overall dataset, respectively \cite{leinster2021}. $\beta$ diversity can be thought of as how distinct a class is, relative to the dataset as a whole. The corresponding quantities for the dataset as a whole are the power means across the classes and are denoted $A$, $B$, and $G$ respectively (``big alpha,'' ``big beta,'' and ``big gamma''). The reciprocal of $\beta$, $\rho=1/\beta$, also has a useful interpretation as how representative a class is of the dataset as a whole; $R$ is the corresponding dataset-wide measure, the power mean of $\rho$ over all classes. Usefully, in LCR, within- and between-class diversities are independent of each other \cite{reeve2016, leinster2021}. (The same is true in the special case that is the Hill framework \cite{jost2007}.) These properties are unique to the the LCR formulation \cite{leinster2021}.

As a terminological note, ``datasets’’ and ``unique elements’’ are examples of what LCR and the broader D-number/ecological literature refer to as ``communities’’ and ``species,’’ respectively; here we use whichever synonym is clearest, given the context.

\subsection{Interpretation}

\begin{figure}[t]
    \begin{center}
    \includegraphics[width = 0.45\textwidth]{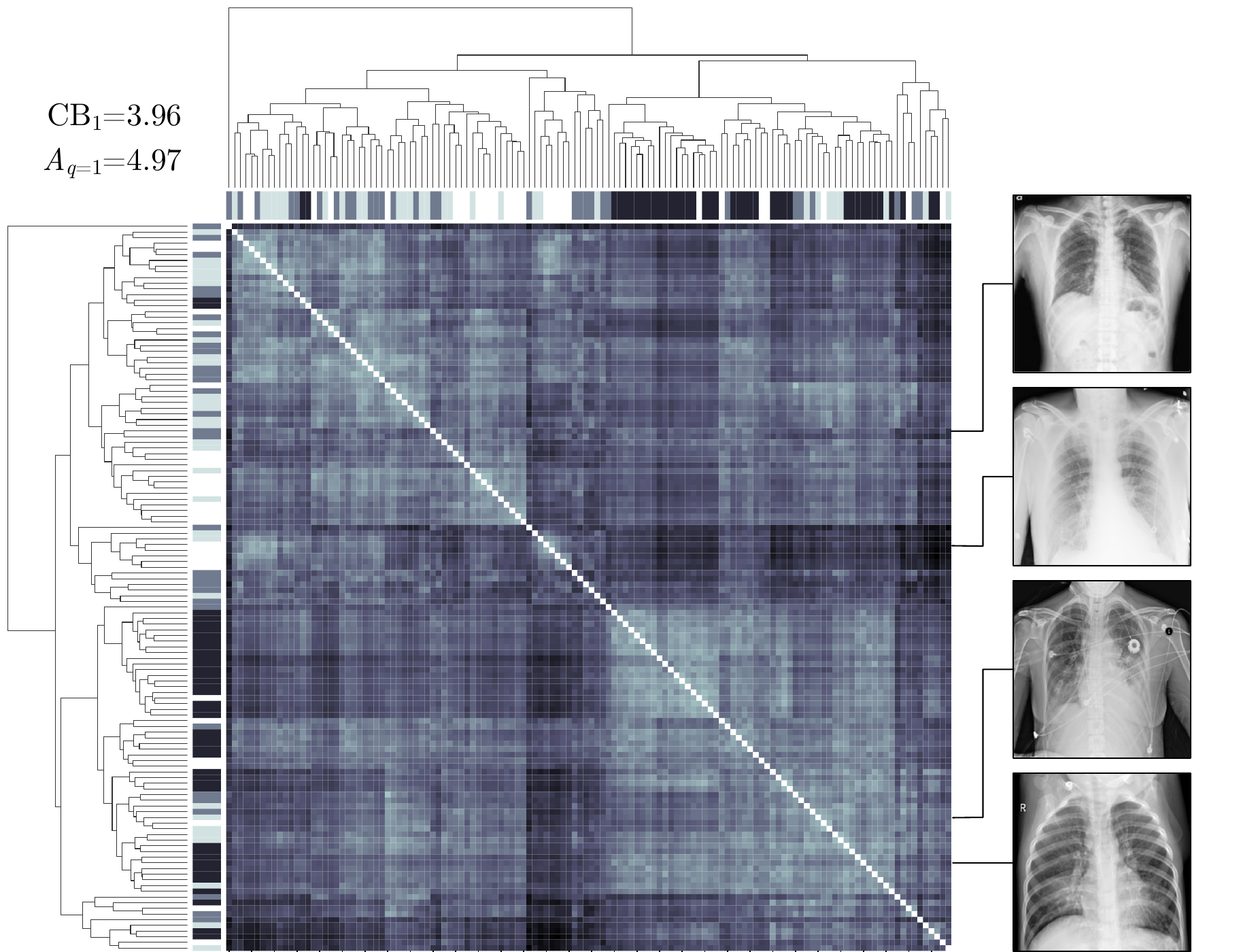}
    \caption{Interpreting the diversity of a dataset by visualizing soft clusters within the similarity matrix. Right: four randomly chosen images from different regions of the clustered heatmap, one from each of the four classes.}
    \label{fig: y}
    \end{center}
\end{figure}

$Z$ can be interpreted as describing a soft clustering of images based on their pairwise similarities (Fig. \ref{fig: y}). In this example, $Z_{ij}=e^{-\text{RMSD}_{ij}}$ for images $i$ and $j$ was calculated for a nearly perfectly class-balanced but otherwise random subset of 128 images from the COVID dataset and clustered using the clustermap method of the seaborn Python package.

The heatmap in Fig. \ref{fig: y} illustrates multiple soft, overlapping, or fuzzy clusters (lighter regions) across the four classes in this small dataset, which are represented by shaded bars along the top and left. $\text{CB}_1$ is the class balance in effective-number form at $q=1$, i.e. the exponential of the class-balance Shannon entropy, a common way of expressing class balance in the literature. $A_1$ is the effective-number form of the power mean of similarity-sensitive Shannon entropies across the four classes, also at $q=1$ (the Shannon analogue). It is interpreted as the effective number of unique image-class pairs in the dataset, after accounting for similarity. The difference between class balance (an effective number of 3.96 classes, meaning all four classes are almost evenly represented) and $A_1$ (4.97 images, meaning this dataset has the same diversity as a dataset with just \textasciitilde{}5 completely distinct/zero-pairwise-similarity images) can be interpreted as the diversity within the four classes, which class balance does not capture. The diversity $D_Z^q(p)$ of the dataset can then be thought of as the effective number of these clusters, with $Z$ accounting for overlap and $q$ describing how to weight the count based on the clusters' relative sizes. 

The ability for $Z_{ij}$ to take multiple/continuous values is what allows for soft/overlapping/fuzzy cluster membership. Which $q$ or $q$s are of interest to the investigator depends on the context, although $q=0$ and $q=\infty$ are natural bounds, with $q=1$ as a sensible midpoint; here we are primarily interested in investigating $q$ that result in $\alpha$, $\rho$, and $\gamma$ and $A$, $R$, and $G$ that correlate with the performance of deep-learning models on image-classification tasks for medical-image datasets.

\section{Methods}\label{sec: methodology}

\begin{table*}[ht]
    \centering
    \caption{Datasets Used in This Study}
    \label{tab:dataset metadata}
     \begin{tabular}
        {|l|rrrrrrr|}
        \hline
        Dataset     & Image size  & Colors & Target               & Modality & No. classes & No. train & No. test \\
        \hline
        PathMNIST   & $ 28 \times 28$  & 3 & Colon histopathology & Microscopy  &  9 &  89,996 &  7,180  \\
        BloodMNIST  & $ 28 \times 28$  & 3 & Blood smears         & Microscopy  &  8 &  11,959 &  3,421  \\
        OrganAMNIST & $ 28 \times 28$  & 1 & Abdomen              & CT, axial   & 11 &  34,581 & 17,778  \\
        OrganCMNIST & $ 28 \times 28$  & 1 & Abdomen              & CT, coronal & 11 &  13,000 &  8,268  \\  
        COVID       & $299 \times 299$ & 1 & Chest                & X-ray       &  4 &  12,288 &  4,096  \\  
        CHD         & $ 80 \times 80$  & 1 & Fetal ECG            & Ultrasound  &  3 &  52,619 &  8,773  \\  
        AVC         & $ 80 \times 60$  & 1 & Adult ECG            & Ultrasound  & 15 & 179,612 & 23,096  \\    
        \hline
    \end{tabular}
\end{table*}

\subsection{Selecting datasets} 

Datasets were selected to represent diverse imaging modalities, provided the following criteria were met: the dataset was large enough to allow extensive subsampling ($\geq$10,000 images), the images were labeled mutually exclusively, and published benchmark accuracy was sufficiently high ($\geq$90\% as reported in the relevant literature).

MedMNIST is a public collection of datasets of different imaging modalities \cite{medmnistv1, medmnistv2}.  Four MedMNIST datasets met the above criteria: BloodMNIST, PathMNIST, OrganAMNIST, and OrganCMNIST. BloodMNIST contains resized, center-cropped, color light-microscope images of single cells from peripheral blood smears from healthy human individuals, with eight classes labeled by cell type: basophil, eosinophil, erythroblast, immature granulocyte, lymphocyte, monocyte, neutrophil, and platelet. PathMNIST is a set of light-microscope images of non-overlapping image patches from hematoxylin-eosin stained histological slides from colorectal cancer cases, with nine classes labeled by tissue type: tumor epithelium, simple stroma, complex stroma, mucosal lymphoid follicle, adipose tissue, debris, normal mucosal glands, duct and vessels, and muscle. OrganAMNIST and OrganCMNIST are, respectively, axial- and coronal-cut computed tomography (CT) images from the Liver Tumor Segmentation Benchmark (LiTS) \cite{lits2023} with 11 classes labeled by organ or bone: bladder, femur-left, femur-right, heart, kidney-left, kidney-right, liver, lung-left, lung-right, pancreas, and spleen.

Three non-MedMNIST datasets were also used, which we refer to as CHD, AVC, and COVID. The CHD dataset consists of fetal echocardiogram images (ultrasounds) from \cite{arnaoutEnsemble2021} labeled by view into three classes: apical four-chamber (A4C), three-vessel (3VV), and nontarget (NT). In ultrasound, one or more clips are acquired per patient; each clip consists of one to several hundred consecutive image frames. Training and test sets were divided by patient identifier and were disjoint from each other \cite{Chinn2021}. AVC is a dataset of echocardiographic images from 267 different studies from adult patients \cite{madani2018}, labeled into 15 classes by view: apical 2 chamber (A2C), apical 3 chamber (A3C), A4C, apical 5 chamber (A5C), parasternal long axis (PSLA), short axis basal (SAXBASAL), short axis mid (SAXMID), subcostal 4 chamber (SUB4C), subcostal IVC (IVC), subcostal aorta (SUBAO), right ventricular inflow (RVinflow), and suprasternal aorta (SUPAO). Finally, COVID is the COVID-19 radiography dataset, which consists of chest X-ray images with four classes labeled by diagnosis: normal, non-COVID lung opacity, COVID-19, and viral pneumonia (presumably not from COVID-19) \cite{chowdhury2020, rahman2021}.

The numbers, sizes, and color depth of the images in each dataset are in Table \ref{tab:dataset metadata}.

\subsection{Creating dataset subsets}

Subsets of different sizes and class balances were created for each of the above datasets, as follows. For each subset size $n$, a large number of partitions of $n$ into a sum of $m$ positive integers was randomly generated, where $m$ is the number of classes in the dataset. These were filtered to ensure that the $i\textsuperscript{th}$ element of the partition was smaller than the number of images in the $i\textsuperscript{th}$ class of the parent training set. For each partition, a corresponding probability distribution was computed by dividing by $n$, and class balance (CB) was measured as the $D_1$ of that probability distribution. Partitions were binned into 25 $D_1$ bins. For $n=$ 64, 128, 256, 384, 512, 768, or 1,024 images, four partitions were chosen at random; for bins with fewer than four partitions, all partitions were selected. 
For each chosen partition $(n_1, n_2, ..., n_m)$, $n_i$ images from class $i$ were chosen uniformly at random from the parent training set. These images formed a subset. This was repeated for each of the datasets.

\subsection{Model training and performance/quality assessment}

A ResNet-18 model \cite{resnet18} was trained on each full dataset (to validate the benchmark performances reported in the literature) and on each subset, using PyTorch 2.1.2 with CUDA 1.21 \cite{pytorch}. The first and last layer were modified to accommodate the differences between datasets in terms of images size and number of classes; the architecture was otherwise unaltered. Because the goal was to study the effect of dataset composition in isolation, no data augmentation was used. 

Each model was trained to a training accuracy of $\geq99$\% or for 1,000 epochs, whichever came first. The trained models were then tested against a test set common to all the subsets from the given parent dataset. In all cases, the test set was disjoint from the parent dataset and all subsets. The quality of the subset (or of the parent dataset) was measured as the performance of the ResNet: the higher the model performance, the higher the quality of the subset (or parent dataset) was considered to be. Performance was measured using the accuracy (ACC), balanced accuracy (BACC), and multiclass area under the receiver-operator characteristic curve (multiclass AUC), all as implemented in PyTorch. Note that in the PyTorch implementation, AUC is the same as multiclass AUROC under default behavior (micro-averaging); ACC is the same as the F1 score, recall, or precision, again with default behavior (micro-averaging); and BACC is the same as recall with macro-averaging.

\subsection{Definition of pairwise image similarity and measurement of diversity features}

For each of the selected subsets, similarity-sensitive diversity features were measured according to the LCR framework using the \textit{greylock} Python package \cite{greylock}. The pairwise similarity $Z_{ij}$ between images $i$ and $j$ was $Z_{ij}=e^{-\text{RMSD}_{ij}}$, where RMSD is the pixel-wise root-mean-squared difference. Note that in practice, many sensible definitions of pairwise image similarity are monotonic with this one.

\subsection{Quality indicators/regression features}

Linear regression was used to measure the adjusted $R^2$ between performance and (the logs of) different feature sets: i.e. size, class balance, and different diversity measures. Each feature is a potential indicator of dataset quality. A total of $30$ features were considered. These included $27$ diversity-related features: $A$, $R$, and $G$ at each of $q=0$, $1$, and $\infty$; minimum $\alpha$, $\rho$, and $\gamma$ at $q=0$ and $1$; maximum $\alpha$, $\rho$, and $\gamma$ at each of $q=0$, $1$, and $\infty$; class balance at $q=1$ and $\infty$; and dataset size. (Recall that $\alpha$, $\rho$, and $\gamma$ are per-class diversity measures; note that minima of $\alpha$, $\rho$, and $\gamma$ measures at $q=\infty$ are the same as/redundant with $A$, $R$, and $G$ at ${q=\infty}$.) These features were chosen as a manageable number of features that nonetheless span the qualitative and quantitative range of information that the LCR framework provides. Note that min and max $\rho$ are just the reciprocals of max and min $\beta$. 

The remaining three features were systematic parameters that differ by dataset but are independent of diversity: image size (the no. of pixels per image), color depth (1 for grayscale images and 3 for color images), and the number of classes (3-15; Table \ref{tab:dataset metadata}). These features were included as a default in all investigations to allow a linear regressor (see next section) the option of using them as normalizing factors. Using the logs allowed the linear regressor to discover scaling relationships, such as class balance scaled by the number of classes. (This is because logging converts ratios into differences, with linear coefficients for each feature.)
\begin{table}[t]
    \centering
    \caption{AUC, Accuracy (ACC), and Balanced Accuracy (BACC) of Benchmark Models for Each Dataset}
    \label{tab: benchmark results}
    \begin{tabular}{|l|rrr|}
        \hline
        Data set & AUC & ACC & BACC \\
        \hline
        PathMNIST   & 0.96 & 0.84 & 0.80 \\
        AVC         & 0.95 & 0.87 & 0.85 \\
        OrganCMNIST & 0.99 & 0.89 & 0.88 \\
        CHD         & 0.98 & 0.92 & 0.88 \\
        OrganAMNIST & 0.99 & 0.92 & 0.91 \\
        BloodMNIST  & 0.99 & 0.93 & 0.92 \\
        COVID       & 0.99 & 0.95 & 0.95 \\
        \hline
    \end{tabular}
\end{table}
\subsection{Assessment of quality indicators/features}

Regressions were performed on each diversity-related feature, each possible pair of diversity-related features, each possible set of three diversity-related features, and other sets of interest. The three systematic parameters above were included in every regression, to allow the regression model to normalize by these features. All regressions were performed using the scikit-learn Python package \cite{scikit-learn}. Adjusted $R^2$ was used in order to account for and correct possible overfitting.

\subsection{Computational requirements}

All computations were done on a single linux machine with 125GB RAM, an 11th-generation Intel Core i9-11900K CPU running at 3.50GHz, and an NVIDIA GA102GL RTX A5000 GPU running at 33MHz. Total runtime to recapitulate the results described in this work on this system was approximately 1 week.

\section{Results}\label{sec: results}

\subsection{Overall goal and strategy}

The goal of this work was to test how dataset quality affects model performance, with quality measured using the diversity measures of the LCR framework and performance measured by BACC and related statistics. This was done in three steps. (1) First, for each of several medical imaging datasets, we created many hundreds of image subsets that differed by size, class balance, and diversity. (2) Second, we trained classifiers on each subset using the same reference model architecture (ResNet-18). (3) And third, we regressed performance against LCR diversity measures, including size and class balance as special cases, to determine how well each measure correlates with model performance. In this way, we sought to identify the best correlates of performance from among the measures tested, and to compare them to the current deep-learning benchmarks of size and class balance.

\subsection{Identifying and validating high-performance datasets}

We identified seven large, well characterized, high-quality datasets suitable for analysis (Table \ref{tab:dataset metadata}). To ensure that results would generalize beyond specific cases, these were chosen to cover a rich variety of tissues and organs and a wide range of imaging modalities: cancer histopathology (light microscopy), blood smears (light microscopy), chest radiographs (X-rays), abdominal CTs, and fetal and adult echocardiograms (ultrasound). These datasets included both color and grayscale images and also varied by image size, supporting the generality of the results.

We confirmed the datasets were high quality by requiring excellent performance on classification tasks to have been documented in the literature by AUC, ACC, and/or BACC. We validated these results by training a reference-architecture (ResNet-18) model on each dataset and confirming strong performance, with median AUC, ACC, and BACC of 0.99, 0.92, and 0.88, respectively despite no data augmentation (Table \ref{tab: benchmark results}). Results from the literature were observed to be slightly better on balance, which was expected because published studies generally used highly specialized and/or fine-tuned models, as opposed to our reference model (whose architecture was held constant except for input and output layers [see Methods] so that the only variable was dataset quality), as well as possibly due to data augmentation. Performance was sufficiently good to ensure a range of performance values to regress against diversity measures, once datasets were divided into subsets.

\subsection{Model performance across subsets of varying size, class balance, and diversity}

\begin{figure*}[ht]
    \centering
        \includegraphics[width=0.9\textwidth]{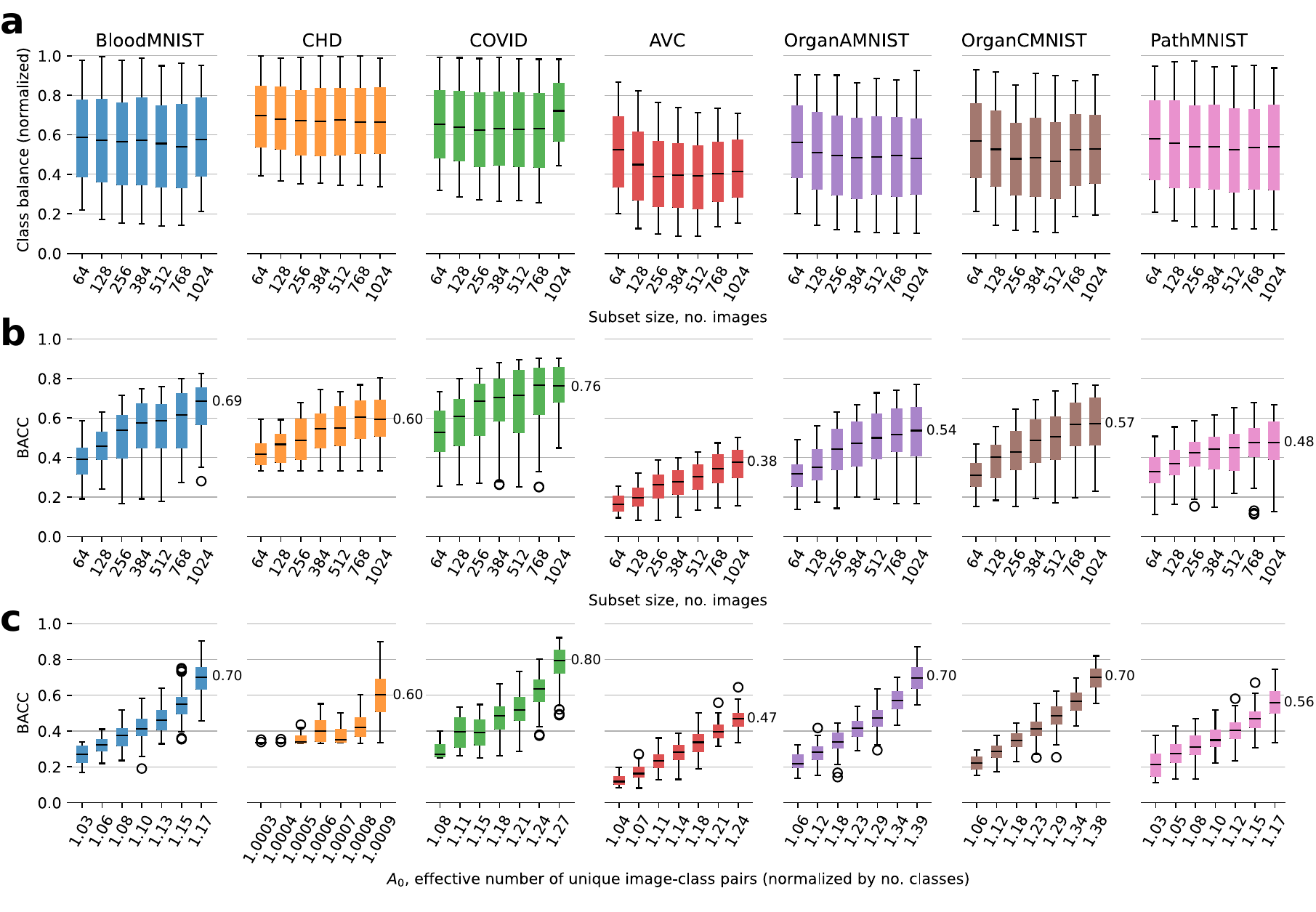}
    \caption{(a) Class balance vs. subset size, (b) BACC vs. subset size, and (c) BACC vs. $A_0$ (normalized by number of classes) for each dataset. Boxes span first and third quartiles; whiskers span an additional 1.5 interquartiles (the matplotlib.pyplot.boxplot default settings). Numbers in (b) and (c) indicate the median BACC for the top bin.}
    \label{fig: class_balance_and_BACC_vs_subset_size}
\end{figure*}

For each dataset, we created hundreds of subsets ranging from 64 to 1,024 images in size that were designed to express the full range of class-balance values that are possible at each size (Fig. \ref{fig: class_balance_and_BACC_vs_subset_size}a and Table \ref{tab:subsets_by_size}). 
\begin{table}[t]
    \centering
    \caption{Number of Subsets of Each Size Chosen from Each Dataset}
    \label{tab:subsets_by_size}
    \begin{tabular}{|l|rrrrrrrrr|}
     \hline
     Dataset Name& 64  & 128 & 256 & 384 & 512 & 768 & 1024 & 2048 & total \\  
     \hline
     PathMNIST   & 88  & 93  & 96  & 96  & 93  & 94  & 96   & 50   & 706   \\
     BloodMNIST  & 90  & 95  & 97  & 100 & 96  & 96  & 88   & 40   & 702   \\
     OrganAMNIST & 80  & 86  & 89  & 86  & 88  & 88  & 86   & 44   & 641   \\
     OrganCMNIST & 81  & 88  & 84  & 87  & 85  & 79  & 80   & 30   & 614   \\
     CHD         & 92  & 96  & 100 & 100 & 100 & 100 & 100  & 50   & 738   \\
     COVID       & 92  & 96  & 100 & 100 & 100 & 100 & 76   & 20   & 684   \\
     AVC         & 73  & 74  & 71  & 70  & 69  & 65  & 61   & 25   & 508   \\
     \hline
     total       & 596 & 628 & 637 & 639 & 631 & 622 & 587  & 259  & 4599  \\
     \hline
    \end{tabular}
\end{table}
The median number of subsets per dataset was 684 (range, 508-738). Note that creating class-balanced subsets from imbalanced datasets gets more difficult as subset size increases, since the number of images in the smallest class puts a limit on class balance; this was the rationale for 1,024 as the maximum subset size. BACC spanned nearly the maximum possible range, from $0.08$ to $0.92\%$, as desired; and rose with dataset size, as expected (Fig. \ref{fig: class_balance_and_BACC_vs_subset_size}b).

\subsection{Alpha diversities as the best correlates of model performance/dataset quality}

\begin{figure*}[ht]
    \centering
        \includegraphics[width=0.9\textwidth]{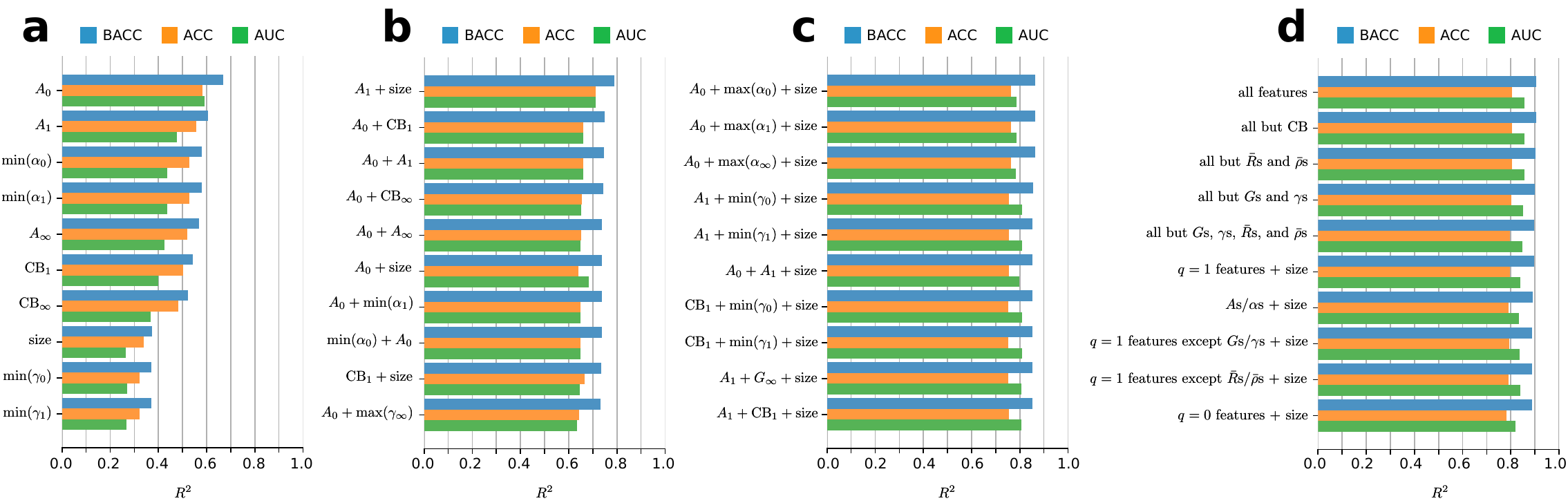}
    \caption{Regression results for (a) top 10 single measures, (b) top 10 pairs of measures, (c) top 10 sets of three measures, and (d) other sets of interest.}
    \label{fig: regression_results}
\end{figure*}

We used regression to assess how size, class balance, and diversity correlate with dataset quality. We tested 30 different features in all: 27 similarity-sensitive diversity measures; two measures of class balance, namely the effective number forms of class Shannon entropy ($\text{CB}_1$) and class Berger-Parker index ($\text{CB}_\infty$); and subset size. Image size (no. pixels), color depth (no. channels), and the number of classes in the dataset were included in all regressions as normalizing factors, since these are constants within each dataset but can differ appreciably between datasets, and since they are unrelated to diversity.

\textbf{Best single correlates.} The five best individual correlates of dataset quality were all alpha diversities; using BACC as the indicator of performance, their mean $R^2$ was 0.600, meaning that on average they explained 60\% of the variance in model performance (Fig. \ref{fig: regression_results}a). In contrast, CBs explained only a little over half the variance in performance ($R^2=$ 0.542 for $\text{CB}_1$ and 0.523 for $\text{CB}_\infty$) and dataset size explained only a little over a third ($R^2=$ 0.373). The best single correlate was $A_0$, which explained two-thirds of the variance in model performance ($R^2=$ 0.670). The second-best was $A_1$, which explained 60.5\% ($R^2=$ 0.605). At the other extreme, the worst $R^2$ was 0.230 (for max $\gamma_\infty$). The pattern for ACC and AUC was similar but with systematically somewhat smaller values (Fig. \ref{fig: regression_results}a). 

$A_0$ was a visibly better correlate of BACC than size (Fig. \ref{fig: class_balance_and_BACC_vs_subset_size}c vs. \ref{fig: class_balance_and_BACC_vs_subset_size}b). Size exhibited plateauing/diminishing returns, as also seen in prior studies such as \cite{althnian2021} (Fig. \ref{fig: class_balance_and_BACC_vs_subset_size}b). Specifically, subsets with the largest $A_0$ were associated with a median of 8\% higher performance than those with the largest size (range, 1-16\%). The largest performance improvements were in the CT, AVC,
and histopathology datasets (Fig. \ref{fig: class_balance_and_BACC_vs_subset_size}b-c). The smallest effects on the medians were in BloodMNIST and CHD, but even in these datasets, the interquartile ranges reflected higher performances for $A_0$ than for size (Fig. \ref{fig: class_balance_and_BACC_vs_subset_size}b-c; compare rightmost colored bars).

\textbf{Best pairs of correlates.} The best pair of correlates was $A_1$ plus dataset size, which together had an $R^2$ for BACC of 0.789 (Fig. \ref{fig: regression_results}b). Nine of the 10 best pairs contained at least one alpha diversity measure; only three contained a CB and only three contained size. The pair of $A_0$ plus $A_1$ was third best at $R^2=$ 0.746, essentially tied with the second-place pair of $A_1$ plus $\text{CB}_1$ ($R^2=$ 0.748). $A_0$ plus $A_\infty$, the only appearance of $A_\infty$ in the top 10 pairs, was fourth at $R^2=$ 0.739. The pair of size plus $\text{CB}_1$ was ninth at $R^2=$ 0.74.

\begin{figure}[t]
    \centering
        \includegraphics[width=0.45\textwidth]{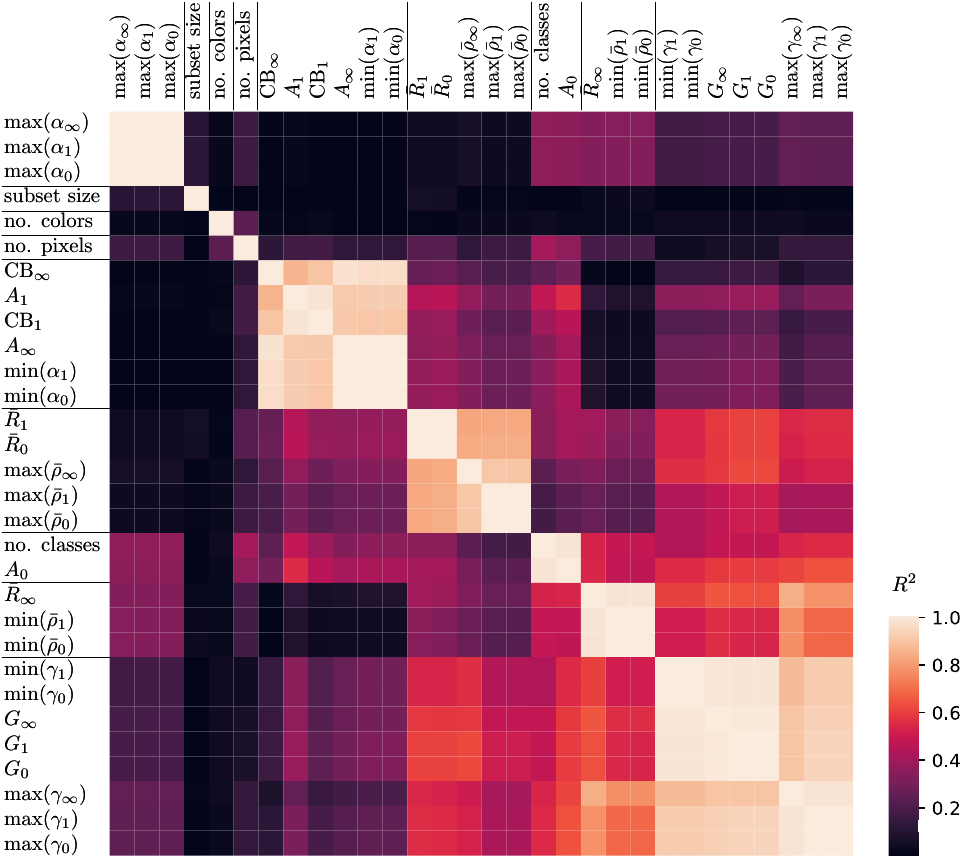}
    \caption{$R^2$ among diversity measures (clustered). Lines delimit highly correlated clusters.}
    \label{fig: correlations}
\end{figure}

\textbf{Other groups of correlates.} The top 10 groups of three correlates all performed similarly, with $R^2$ values of 0.840-0.851 for BACC (Fig. \ref{fig: regression_results}c-d). Each included size, and eight of the 10 included $A_0$ or $A_1$. The only two groups that did not contain an $A$ included $\text{CB}_1$ and either min $\gamma_0$ or min $\gamma_1$. Pairwise correlations among the features are shown in Fig. \ref{fig: correlations}. Collectively, the eight alpha diversities tested---$A_0$, $A_1$, $A_\infty$, min $\alpha_0$, min $\alpha_1$, max $\alpha_0$, max $\alpha_1$, and max $\alpha_\infty$---explained 78.6\% of the variance in performance. This was appreciably better than the eight gamma or rho diversities (62.4\% and 59.0\%, respectively) or the class balances (54.5\%; almost identical to the 54.2\% for $\text{CB}_1$ by itself). For further comparison, recall from above that size on its own explained just 37.3\%. Finally, the combination of all 27 diversity-related features, including size and class balance, explained 90.6\% of the variance in model performance ($R^2=$ 0.906). Fig. \ref{fig: regression_results}d shows additional feature sets of interest.

\textbf{Correlations among diversity measures.} High correlations were observed between many pairs of diversity measures (Fig. \ref{fig: correlations}). These included between $A_0$ and $A_1$, between these $A$s and $G$s, and between these $A$s and the CBs.

\section{Discussion}\label{sec: discussion}

Although the primary approach to maximizing dataset diversity in imaging datasets, as an indicator or guarantor of quality, is to maximize dataset size and class balance, neither of these measures accounts for the diversity of the images themselves (Fig. \ref{fig: fig_diversity}). In the present work, we show that dataset diversity measures that account for the similarities and differences among images outperform size and class balance as indicators of dataset quality, potentially offering ways to improve model performance without additional compute power or architectural improvements, by optimizing for $A$ and/or other measures in the comprehensive LCR framework (which includes size and class balance as special cases). 

We find that the alpha-diversity measures $A_0$ and $A_1$ are better correlates of model performance than size or class balance, and are thereby better measures of dataset quality. Specifically, $A_0$ explained about two-thirds of the variance in BACC, or nearly twice as much variance as dataset size. Meanwhile, $\text{CB}_1$ explained a little over half the variance. On average, datasets with the largest $A_0$ had a median 8\% higher accuracy than those with the largest size (Fig. \ref{fig: class_balance_and_BACC_vs_subset_size}b-c). We consider these meaningful improvements.

\textbf{$A_0$ as the top correlate of dataset quality.} Conceptually, $A_q$ is the effective number of unique image-class pairs in the dataset, down-weighting less-frequent pairs proportional to the given $q$, given the similarities and differences among the images. They measure the alpha diversity of a dataset: they are averages of the alpha diversities of the labeled classes (taking the similarities and differences among images into account). $A_0$ is simply the arithmetic mean of the $\alpha_0$ diversities of the classes; $A_1$ is reminiscent of Shannon entropy in that it down-weights the contributions of less diverse classes by an amount given by $q=1$. 

We found that $A_0$, the best correlate of model performance, appreciably outperformed $A_1$: a 6.5\% absolute and 10.7\% relative improvement (for BACC). However, adding size reversed the relationship: $A_1$ plus size outperformed $A_0$ plus size, albeit by a smaller amount (5.1\% absolute/6.9\% relative improvement), consistent with $A_1$ being more independent of size. We note that $A_0$ and $A_1$ were highly correlated with each other. Because we did not explicitly investigate methods for maximizing the different diversity measures, it is not clear from the present work whether one would have to choose one or the other $A$ to maximize, or whether both of them (and/or other diversity measures) could be maximized simultaneously. Similarly, it is not obvious when it might be desirable to maximize one diversity measure at the expense of another. These are questions for future work.

\textbf{Alphas vs. other diversity measures.} It is interesting that $A_0$ and $A_1$ both outperform $G$ by a considerable margin. Indeed $A_0$ outperforms all the gamma-diversity measures combined (67.0\% vs. 62.4\% for all gammas combined). $G$ differs from $A$ in being the effective number of unique images in the dataset irrespective of class divisions. Thus, our results support the intuition that high-quality datasets should have diverse classes, not just diverse images. This is sensible: one would expect models to have an easier time learning determinative class invariants when each class represents diverse images. This understanding explains the much weaker performance of $A_\infty$ (which scored about as well as class balance): the $q=\infty$ in $A_\infty$ means that this $A$ focuses on just the single most diverse class. Our results show that the diversity of all classes matters, not just that of the most diverse class. 

Of the dataset-level diversity measures, the ones that showed the lowest correlation with performance were the $\bar{\rho}$s (28.3-35.0\%) and $R$s (23.9-27.8\%). These were absent from the top-correlating sets of features. Interestingly, together all $\bar{\rho}$s and $R$s explained 59.0\% of the variance in BACC (better than CB). Recall that $\bar{\rho}_q$ measures how representative each class is of the overall dataset; as $q$ increases, $\bar{\rho}_q$ focuses increasingly on the representativeness of the largest similarity block. $R_q$ is the power mean of $\bar{\rho}_q$ across the classes, interpreted as the amount of information gained about the class, by knowing the pixel values. (For example, for $q=1$, $-\log_2\bar{R}$ is the number of bits of information gained, which is $\sim$0 when $\bar{R}\sim$1, and increases as $\bar{R}$ decreases.) The low correlations of individual $\bar{\rho}$s and $R$s indicate performance does not depend much on how representative the classes are of the dataset as a whole. Their much better performance as a set suggests some complementarity to each other. Exploring this further is left for future work.

\textbf{Correlations among diversity measures.} The strong correlations observed among several diversity measures are a consequence of the relationship $A=\text{CB}\times G\times \bar{R}$ \cite{leinster2021} (which holds precisely at $q=1$ and approximately, in practice, for many other $q$) and the generally high similarity between pairs of images according to the similarity measure we chose. The empirical observation that $A$ correlates well with CB means that $G\times \bar{R}$ is fairly constant across the datasets in this study. Indeed, we observed that both $G$ and $\bar{R}$ are fairly close to 1 (with $G$ between 1 and 1.6). Recall that $G$ is the effective number of unique images; it being close to 1 means the effective number of images in these datasets is always nearly 1, again a consequence of the scale of the similarity measure used in this study. Meanwhile, $\bar{R}$ being close to 1 indicates that knowing the pixel values adds little information regarding the class. This could be because of the large number of potential pixel distributions, or the relatively small number of classes in some of the datasets.

\textbf{Limitations and future directions.} To our knowledge, the present work is the first systematic, quantitative demonstration of the relative value of size, class balance, and the LCR measures (a.k.a. similarity-sensitive Hill numbers or effective-number forms of similarity-sensitive R\`enyi entropies) in machine learning, demonstrating these measures combined explain over 90\% of the variance in BACC. However, there are several limitations and questions to be addressed in future work, in addition to those mentioned in the foregoing discussion. 

First, we have considered the quality of the training set but not the validation and/or test sets. Prior work by others suggests that performance will be higher the more representative the validation and test sets are of the training set, but this is yet to investigate through the framework of LCR. Second, it will be interesting to explore the effect of different image similarity measures. Here we tested a simple RMSD-based measure, which is monotonic with other measures such as VAE embeddings \cite{chinn2023a}. A more complete investigation of the effects of different similarity measures may prove insightful. Third, in order to facilitate normalization by image size, color depth, and number of classes, we used features' logarithms for our regressor. Whether other potential relationships besides the $\log$ might be more informative or more natural remains to be seen. Perhaps a different transform would account for the $\sim$10\% of variance unaccounted for by the subset of LCR measures we investigated. Fourth, it will be interesting to test how well our approach can predict which of two datasets will result in a higher-performing model, as a potential aid to dataset creation/curation. And fifth, while we have tested a variety of medical imaging datasets, it will be interesting to test how our results generalize to other imaging datasets and to non-imaging datasets (e.g. text, tabular data). 

Overall, the introduction of the rich framework of similarity-sensitive diversity to machine learning is expected to open fruitful directions of future study.

\hfill

\bibliographystyle{IEEEtran}
\bibliography{ref}

%\newpage

\vskip -2\baselineskip plus -1fil

\begin{IEEEbiographynophoto}{Josiah Couch}
recieved B.S. degrees in physics and mathematics respectively from Oklahoma State University in 2013, and received a Ph.D. degree in physics from the University of Texas at Austin in 2021. From 2021 to 2022 he worked as a Postdoc in the Computer Science department at Boston College. He currently works as a Postdoctoral Research Fellow at Beth Israel Deaconess Medical Center in Boston, MA on applications of machine learning to healthcare. He has worked on topics related to quantum gravity, quantum information theory, the AdS/CFT correspondence, and random graphs. His current research interests include energy based models and information theory. Dr. Couch is a member of the American Physical Society.
\end{IEEEbiographynophoto}

\vskip -2\baselineskip plus -1fil

\begin{IEEEbiographynophoto}{Rima Arnaout} received an S.B. degree in biology from the Massachusetts Institute of Technology in 2002 and an M.D. degree from Harvard Medical School in 2007. She is Associate Professor of Medicine at the University of California, San Francisco (UCSF) in San Francisco, California, USA. She is also faculty in the Baker Institute and the UCSF-UC Berkeley Joint Program for Computational Precision Health. Her research interests are computational cardiology, cardiac imaging, medical imaging, machine learning, artificial intelligence, and computer vision. Dr. Arnaout is a Fellow of the American Institute for Medical and Biological Engineering and an Emerging Leader at the National Academy of Medicine.
\end{IEEEbiographynophoto}

\vskip -2\baselineskip plus -1fil

\begin{IEEEbiographynophoto}{Ramy Arnaout} received an S.B. degree in biology from the Massachusetts Institute of Technology in 1997, a D.Phil. (Ph.D.) degree in Mathematical Biology (Zoology) from the University of Oxford in 1999, and an M.D. degree with honors from Harvard Medical School in 2003. He is Associate Professor of Pathology at Harvard Medical School and Associate Director of the Clinical Microbiology Laboratories at the Beth Israel Deaconess Medical Center, both in Boston, MA, USA. His past and present research interests include entropy, diversity, immunomics, metagenomics, computational pathology, informatics, artificial intelligence, and machine learning. Dr. Arnaout is a member of the College of American Pathologists.\end{IEEEbiographynophoto}

\end{document}